\documentclass[conference]{IEEEtran}
\IEEEoverridecommandlockouts

\usepackage{cite}
\usepackage{amsmath,amssymb,amsfonts}
\usepackage{algorithmic}
\usepackage{graphicx}
\usepackage{textcomp}
\usepackage{xcolor}
\usepackage[a4paper, total={184mm,239mm}]{geometry}
\def\BibTeX{{\rm B\kern-.05em{\sc i\kern-.025em b}\kern-.08em
		T\kern-.1667em\lower.7ex\hbox{E}\kern-.125emX}}

\usepackage{caption}
\usepackage{graphics}
\usepackage{subcaption}
\usepackage{booktabs}
\usepackage{tabularx}
\usepackage{enumitem}
\usepackage{booktabs}
\usepackage{multirow}
\usepackage{siunitx}
\usepackage{balance}
\usepackage{mathtools}
\usepackage[para,online,flushleft]{threeparttable}
\usepackage[linesnumbered, noend, algoruled,boxed,lined]{algorithm2e}
\usepackage{soul}
\usepackage{hyperref}
\hypersetup{%
	pdfborder = {0 0 0}
}

\usepackage[absolute,overlay]{textpos}

\DeclareMathOperator{\MLP}{MLP}
\DeclareMathOperator{\AGG}{AGG}
\DeclareMathOperator{\HD}{HD}
\DeclareMathOperator{\ADRS}{ADRS}

\DeclareRobustCommand*{\IEEEauthorrefmark}[1]{%
	\raisebox{0pt}[0pt][0pt]{\textsuperscript{\footnotesize #1}}%
}

\makeatletter 
\def\footnoterule{\kern-3\p@
	\hrule \@width 2in \kern 2.6\p@} 
\makeatother

\begin{document}
	\begin{textblock*}{21cm}(1cm,27cm) 
		Preprint of the paper to be published in \textit{Design, Automation and Test in Europe Conference and Exhibition (DATE 2022)}
	\end{textblock*}

	\begin{textblock*}{21cm}(6cm,27.5cm) 
		Open source at: \url{https://github.com/zlinaf/PowerGear}
	\end{textblock*}
	
	\title{PowerGear: Early-Stage Power Estimation in FPGA HLS via Heterogeneous Edge-Centric GNNs\vspace{-0.8ex}\\}
	
	\author{\IEEEauthorblockN{Zhe Lin\IEEEauthorrefmark{1}, Zike Yuan\IEEEauthorrefmark{2}\textsuperscript{\dag}, Jieru Zhao\IEEEauthorrefmark{3}, Wei Zhang\IEEEauthorrefmark{4}, Hui Wang\IEEEauthorrefmark{1} and Yonghong Tian\IEEEauthorrefmark{1,5}\textsuperscript{*}}
		\IEEEauthorblockA{\IEEEauthorrefmark{1}Peng Cheng Laboratory, China
			\IEEEauthorrefmark{2}The University of Auckland, New Zealand
			\IEEEauthorrefmark{3}Shanghai Jiao Tong University, China\\
			\IEEEauthorrefmark{4}The Hong Kong University of Science and Technology, Hong Kong, China
			\IEEEauthorrefmark{5}Peking University, China}
		\{linzh01,wangh06,tianyh\}@pcl.ac.cn, zyua138@aucklanduni.ac.nz, zhao-jieru@sjtu.edu.cn, wei.zhang@ust.hk\vspace{-4.3ex}}
	
	\maketitle
	
	\begingroup\renewcommand\thefootnote{\dag}
	\footnotetext{Work done during Zike Yuan's internship at Peng Cheng Laboratory.}
	\begingroup\renewcommand\thefootnote{*}
	\footnotetext{Corresponding author: Yonghong Tian (tianyh@pcl.ac.cn).}

	\setlength{\abovedisplayskip}{2pt}
	\setlength{\belowdisplayskip}{2pt}
	
	\setul{2pt}{.4pt}
	
	\begin{abstract}
		Power estimation is the basis of many hardware optimization strategies. However, it is still challenging to offer accurate power estimation at an early stage such as high-level synthesis (HLS). In this paper, we propose \textit{PowerGear}, a \ul{g}raph-l\ul{ear}ning-assisted \ul{power} estimation approach for FPGA HLS, which features high accuracy, efficiency and transferability. PowerGear comprises two main components: a graph construction flow and a customized graph neural network (GNN) model. Specifically, in the graph construction flow, we introduce buffer insertion, datapath merging, graph trimming and feature annotation techniques to transform HLS designs into graph-structured data, which encode both intra-operation micro-architectures and inter-operation interconnects annotated with switching activities. Furthermore, we propose a novel power-aware heterogeneous edge-centric GNN model which effectively learns heterogeneous edge semantics and structural properties of the constructed graphs via edge-centric neighborhood aggregation, and fits the formulation of dynamic power. Compared with on-board measurement, PowerGear estimates total and dynamic power for new HLS designs with errors of 3.60\% and 8.81\%, respectively, which outperforms the prior arts in research and the commercial product Vivado. In addition, PowerGear demonstrates a speedup of 4$\times$ over Vivado power estimator. Finally, we present a case study in which PowerGear is exploited to facilitate design space exploration for FPGA HLS, leading to a performance gain of up to 11.2\%, compared with methods using state-of-the-art predictive models.
	\end{abstract}
	
	

\section{Introduction}
\setul{2.5pt}{.4pt}
Power efficiency has emerged as one of the first-order constraints for hardware systems such as field-programmable gate arrays (FPGAs), and the design optimization with regard to power efficiency usually necessitates the knowledge of power consumption. However, the power evaluation flow for FPGA designs induces large overheads of design turnaround time. In general, accurate FPGA power estimation requires to obtain the signal activities of critical components and I/O ports via vector-based gate-level simulation, and a set of physical component measurements obtained through the register-transfer level (RTL)-based FPGA implementation flow, including synthesis, placement and routing, etc. With the low-level hardware details disclosed after conducting the above steps, analytical models~\cite{liu94} can be used to deduce signal activities for internal components and then infer power consumption. Therein, the cycle-accurate gate-level simulation and the FPGA implementation flow with NP-complete problems give rise to a large amount of runtime. Overall, in a power-oriented hardware optimization loop, designers have to repeatedly perform the above power evaluation steps while refining the design architecture until power closure is achieved, which incurs long development time and high labor cost for every single design.

To speed up hardware power estimation, great efforts have been made. As for RTL, some works~\cite{zhe18,zhou19} propose to collect a small set of signal activities through RTL simulation and then use them as input features to construct learning-based models for power inference, including decision trees, ensemble models, and convolutional neural networks. These works expedite the power estimation process by replacing the inefficient FPGA implementation flow with efficient modeling strategies. Some other works such as~\cite{zhang20} seek to build models to predict gate-level internal signal activities using I/O and register activities from RTL simulation, and leverage the commercial power analysis tools to conduct power estimation after physical implementation, avoiding the tedious vector-based gate-level simulation step. These works showcase gains in runtime efficiency by skipping either the RTL-based implementation flow or gate-level simulation, while the remaining steps are still time-consuming. 

Some prior works~\cite{zuo15,lee15,lin20} aim at achieving total power estimation at a higher abstraction level, i.e., high-level synthesis (HLS)~\cite{cous09}, which greatly accelerates power estimation by dispensing with the low-level simulation and implementation steps in power inference. Some works~\cite{zuo15,lee15} extract activity features during C-level program execution, and either develop analytical power models for specific types of functions~\cite{zuo15} or utilize machine learning models~\cite{lee15} to learn power characteristics of downstream FPGA implementation. These works are design-specific and the built models are not applicable to new designs of interest. A recent work HL-Pow~\cite{lin20} adopts histograms as a way of feature alignment over different designs, which allows power inference for unseen cases. This is accomplished by encoding the activities of each type of HLS operations into a histogram individually, concatenating histograms as overall design features, and then training models to infer power.

Nevertheless, these prior arts on HLS power estimation mainly carry out feature engineering on individual operations, neglecting the impact of interconnects between different operations and the switching activities associated with interconnects, even though interconnects have been proven to significantly contribute to power dissipation~\cite{zhong02}, especially dynamic power. In order to improve early-stage power estimation for FPGA HLS, we argue that it is vital to take into account the impact of interconnects in power modeling. To this end, we leverage graph neural networks (GNNs) in an effort to jointly learn micro-architectures of HLS operations and interconnects with switching activities. Specifically, we present a graph construction flow to transform the HLS-based hardware designs into graph-structured data, wherein operations are cast as nodes with features indicative of micro-architectures, interconnects are projected as edges with relation types, and netlist activities are encoded as edge features. On this basis, we devise a novel \underline{h}eterogeneous \underline{e}dge-\underline{c}entric GNN model for power modeling, called \textit{HEC-GNN}. Different from general-purpose GNNs that primarily work on node features, HEC-GNN is enhanced with the capability to exploit informative heterogeneous edge semantics and structural properties via the novel edge-centric aggregation scheme, and is aware of power via adaptively approximating the formation of dynamic power.

Putting it all together, we propose a \ul{g}raph-l\ul{ear}ning-assisted \ul{power} modeling approach for FPGA HLS, named \textit{PowerGear}, which distinguishes itself from the prior arts with the following key features: 1) \textbf{accuracy}: PowerGear demonstrates high accuracy for estimating dynamic power besides total power, which has not been jointly studied in prior related works~\cite{zuo15,lee15,lin20} on HLS; 2) \textbf{efficiency}: compared with RTL-based works~\cite{zhe18,zhou19,zhang20} and Vivado integrated power estimator~\cite{vpwr}, PowerGear gains considerable improvement in the turnaround time of power modeling; and 3) \textbf{transferability}: in contrast to the task-dependent modeling frameworks~\cite{zuo15,lee15,zhe18,zhou19}, PowerGear can generalize to previously unseen designs without requiring model retraining. To the best of our knowledge, this is the first attempt to apply customized GNNs for early-stage power estimation in HLS with both total and dynamic power predicted accurately. In all, our major contributions are listed as follows:
\begin{itemize}
	\item We introduce a graph construction flow to convert the FPGA HLS designs into graph data that preserve both intra-operation and inter-operation power-related features.
	\item We propose a novel power-aware GNN model, HEC-GNN, which fully mines rich heterogeneous edge semantics and structural properties via the edge-centric neighborhood aggregation, and subtly fits the dynamic power formulation.
	\item For the first time, we investigate dynamic power estimation for FPGA HLS in addition to total power, and demonstrate how our power estimation approach benefits power-efficient design space exploration in HLS.
\end{itemize}

\section{Preliminaries}
The FPGA power consumption can be broken down into two components: static power and dynamic power consumption. The static power consumption mainly stems from the leakage currents passing through the transistors whenever the devices are powered up, independently of the workloads running in real time.
As for earlier FPGA series, the static power is constant regardless of the implemented designs. However, recent FPGA products, such as Xilinx Ultrascale FPGA, have enabled automatic power gating~\cite{vpwrtech} on unused hard blocks such as logic units and memories, making the static power dependent upon the design scale and utilized resource types.

The dynamic power consumption, on the other hand, is caused by signal switching activities, i.e., transistor switches or register value changes that repeatedly charge and discharge interconnect capacitance. In contrast to static power, dynamic power reflects the runtime workloads driven by specific data rates and data characteristics. The overall dynamic power consumption is the sum of power consumption for each single netlist component triggering signal toggling, which is
\begin{equation}
	\label{eq:dyn}
	\small
	P_{dyn} = \sum_{i\in I}{\alpha_iC_iV^2f},
\end{equation}
where $\alpha$ is the signal switching activity, $C$ is the interconnect capacitance, $V$ is the supply voltage, $f$ is the operating frequency and $i$ is an interconnect of the whole set $I$. For a specific FPGA, the supply voltage $V$ is fixed, and $C$ for each $i$ is decided by the FPGA placement and routing algorithms. Hence, given a target design on a specific FPGA, dynamic power is largely determined by runtime workloads that give rise to different signal switching activities and operating frequencies. 

\section{PowerGear Methodology}
\begin{figure}[t]
	\begin{center}
		\includegraphics[width=0.8\linewidth]{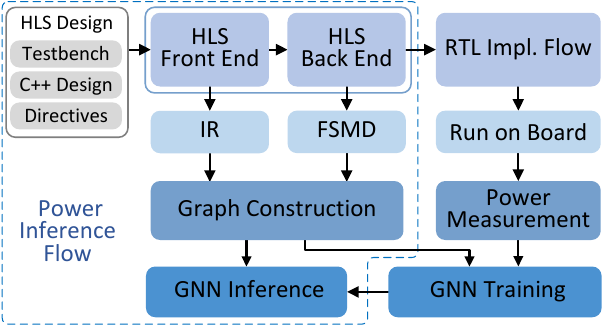}
		\caption{PowerGear overview: in the training stage, graph-structured samples are constructed using HLS results, ground truth power values are collected from real measurement on board after RTL implementation, and a transferable GNN model is trained; in the inference stage, graph samples for new designs are generated after HLS runs and the trained GNN model is seamlessly employed for power prediction.}
		\label{fig:pgview}
	\end{center}
	\vspace{-7mm}
\end{figure}

The overview of our proposed approach, PowerGear, is shown in Fig.~\ref{fig:pgview}. To start with, we investigate an automated tool flow to capture the power-related information in the HLS designs and generate graph-structured data samples encapsulating both individual and contextual information of various types of design components. Therein, the graph construction is solely based on HLS results, i.e., the intermediate representation (IR) from HLS front-end compilation, and the finite state machine with datapath (FSMD) from HLS back-end optimization. The ground truth power value of each sample for model building is collected by real measurement on FPGA after performing the RTL implementation flow. In the training stage, we develop a novel directed, heterogeneous and edge-centric GNN model to effectively learn the principles behind FPGA power consumption, using the graph-structured features and the corresponding ground truth power values. In the power inference stage, new designs of interest only need to pass through the HLS flow, be converted into graph data in the same way as the training stage, and be fed into the trained GNN model to infer power directly, skipping the RTL implementation and model construction steps in the training stage. In the following parts of this section, we separately describe our proposed automated graph construction flow and GNN-assisted power modeling strategy.

\vspace{-1mm}
\subsection{Graph Construction Flow}
\label{subsec:gcf}
In the graph construction flow, we propose to generate graph-structured samples for FPGA HLS designs in order to make use of GNNs for power learning. HLS tool flows usually provide IR and FSMD information to recover dataflow graphs (DFGs)~\cite{ustun20}. Indeed, the DFGs have already presented graph-like data: each DFG node is associated with an IR operation which defines its micro-architecture, and the interconnect between two operations forms a DFG edge. Directly using the HLS DFGs for power modeling, however, is problematic, because primitive DFGs tend to be large-scale but convey limited power-related details, harming both model efficiency and efficacy. To tackle this problem, we introduce four optimization strategies on the HLS DFGs to retrieve and retain from the DFG nodes the hardware components that have significant influence on power consumption, and augment the interconnects, i.e, DFG edges, with power information, which is depicted in Fig.~\ref{fig:gf}.
\begin{figure}[t]
	\begin{center}
		\includegraphics[width=0.98\linewidth]{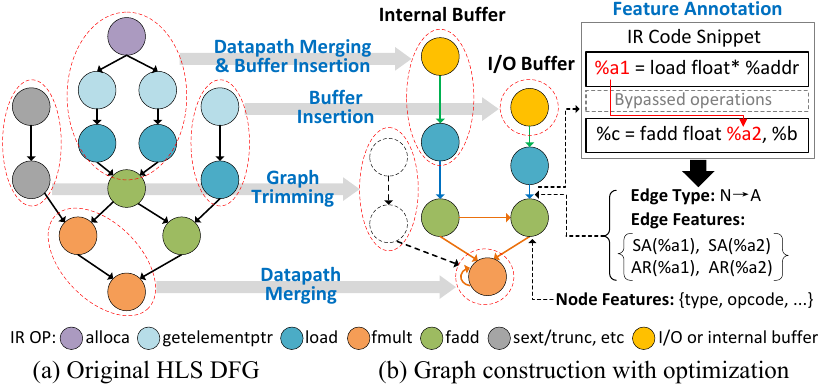}
		\vspace{-1mm}
		\caption{Graph construction flow with HLS DFG optimization.}
		\label{fig:gf}
	\end{center}
	\vspace{-9mm}
\end{figure}

\textbf{Buffer insertion.} In the DFGs, buffer components, including internal and I/O buffers, have not been declared explicitly. Nevertheless, memory activity is a critical source of power consumption. Hence, we seek to instrument buffers in the DFGs. We observe that memory elements can be inferred from DFG nodes with IR opcodes \textit{alloca} and \textit{getelementptr} followed by \textit{load} or \textit{store}. Specifically, we perform pattern matching of the above memory-related DFG nodes, insert buffer nodes in DFGs, connect buffer nodes to downstream nodes, and annotate buffer nodes with memory resource utilization.

\textbf{Datapath merging.} The DFGs may contain plenty of identical node chains from a source node to a sink node. For instance, loading data from a specific buffer and storing back to another one in different loop execution may cause different IR operations to be instantiated, even though these operations are highly similar. This leads to multiple duplicate DFG node chains generated between two DFG nodes, which deviates from the exact hardware realization. To restore the real hardware implementation from the DFGs, we perform datapath merging to fuse identical DFG node chains between two nodes. Furthermore, we try to account for resource sharing between different FSM stages in RTL, and to this end, we merge the DFG nodes utilizing the same set of hardware resources.

\textbf{Graph trimming.} The DFGs encompass paths that are not computationally intensive. We bypass DFG nodes that contribute little to arithmetic computation and also produce trivial hardware entities, e.g., bit truncation and signed extension. This helps to improve the modeling efficiency by reducing the scale of the generated graph samples, and suppress noise by making the model focus more on crucial arithmetic-intensive datapaths.

\textbf{Feature annotation.} The DFGs do not reflect any signal toggling intrinsically. As the signal activity is the trigger of dynamic power consumption, we propose to integrate signal activities in interconnects, i.e., edges of graph samples for power learning. First, we annotate the signals across each edge of interest. That is, we identify the dataflow variables (operands or results of IR operations) transferred through the edges from the source nodes to the sink nodes. Then, we instrument detection probes to trace the values of variables of interest in the IR level. Next, we link together the testbenches with input stimuli, the instrumented IR and the function entities of detection probes following the method of~\cite{lin20}. Subsequently, we compile them into a single executable, and run the executable to acquire traced variable values in different execution cycles. Finally, we compute edge switching activities $SA_{edge}$ by considering data injecting in and going out of edges individually. This gives
\begin{equation}
	\label{eq:sa}
	\scriptsize
	SA_{edge} = \bigg\{ \frac{\sum_{i=1}^{N_{\boldsymbol{v}}^{dir}}\HD\big(\boldsymbol{v}_{dir}(i), \boldsymbol{v}_{dir}(i-1)\big)}{L} \ \bigg|\ dir \in \{src, snk\} \bigg\},
\end{equation}
where $\boldsymbol{v}$ is the set of bit vectors of variable values that pass through the edge and change at different execution cycles, $dir$ is the dataflow direction that separates $\boldsymbol{v}$ produced by source nodes ($src$) from $\boldsymbol{v}$ utilized by sink nodes ($snk$) of the edge, $N_{\boldsymbol{v}}^{dir}$ is the number of execution cycles that cause the change of $\boldsymbol{v}$ with $dir$, $L$ is the latency of the design, and $\HD(\cdot)$ is the Hamming distance computation which counts the number of variable bits that are different in two execution cycles. Besides $SA_{edge}$, we also extract the activation rate for each edge, which is defined as 
\begin{equation}
	\label{eq:ar}
	\scriptsize
	AR_{edge} = \bigg\{ \frac{N_{\boldsymbol{v}}^{dir}}{L} \ \bigg|\ dir \in \{src, snk\} \bigg\}.
\end{equation}

Putting it all together, we construct four-dimensional edge features comprising the switching activities and the activation rates of both the source and sink variables through the edge. We further classify the DFG nodes into two categories: arithmetic (A) and non-arithmetic (N) nodes. Correspondingly, the edges are annotated with four types of source-to-sink relations: A$\rightarrow$N, A$\rightarrow$A, N$\rightarrow$A, and N$\rightarrow$N. As for nodes, we use the IR operation type, IR operation opcode, overall activation rate, input, output and overall switching activities as node features. The IR operation type and opcode are categorical features with one-hot encoding while the others are numeric features.

\vspace{-0.7mm}
\subsection{Heterogeneous Edge-Centric GNN Model}
The graphs generated by our proposed construction flow exhibit the following characteristics: 1) \textbf{heterogeneity}: the graph edges have different relation types, indicating whether the corresponding datapaths comprise arithmetic operations; 2) \textbf{edge expressivity}: signal switching activities, the major contributors to dynamic power as shown in Eq.~\ref{eq:dyn}, are intactly encoded in the edges, which demonstrates that the edge features have higher expressive capability than node features in terms of power consumption; 3) \textbf{directionality}: the edges are directed, meaning that data can only pass through an edge from the source node to the sink node, while the opposite is infeasible.

The mainstream GNN models~\cite{kipf17,hamilton17} mostly pay attention to node feature aggregation. Some recent works~\cite{hu20,morris19} also support using edge features as the supplement to node features, whereas node features still play a dominant role in message passing. These existing node-centric GNN models, however, are ineffective in capturing heterogeneous edge semantics that are essential in the context of power estimation. In light of this, we develop a power-aware \underline{h}eterogeneous \underline{e}dge-\underline{c}entric GNN model, named \textit{HEC-GNN}, which makes full use of edges in neighborhood aggregation to benefit power modeling.

\setlength{\belowdisplayskip}{1pt}
\textbf{HEC-GNN convolutional layer}. Each graph sample built with our construction flow can be represented as $G = (\mathcal{V}, \mathcal{E}, \mathcal{R})$, where $\mathcal{V}$, $\mathcal{E}$ and $\mathcal{R}$ denote the set of graph nodes, edges and relation types, respectively. Formally, our proposed HEC-GNN collects information from neighbors and updates node embeddings at the $k$-th convolutional layer as
\begin{equation}
	\label{eq:layer}
	\footnotesize
	\boldsymbol{h}^{(k)}_v = \sigma\left(\boldsymbol{W}_\mathcal{V}^{(k)} \boldsymbol{h}_v^{(k-1)} + \AGG^{(k)}(\{ \boldsymbol{e}_{u,v,r}| r \in \mathcal{R}, u \in \mathcal{N}_v^r\})\right),
\end{equation}
where $\boldsymbol{h}^{(k)}_v$ is the embedding vector of node $v\in\mathcal{V}$, $\boldsymbol{e}_{u,v,r}$ is the edge feature vector of the directed edge from node $u$ to node $v$ with relation type $r\in\mathcal{R}$ and $(u,v,r)\in\mathcal{E}$, $\mathcal{N}^r_v$ is the set of nodes with the successor being node $v$ and the edge relation type being $r$, $\boldsymbol{W}_\mathcal{V}^{(k)}$ is the learnable weight matrix for updating node embeddings from the last layer, $\sigma$ is the ReLU activation function, and $\AGG^{(k)}$ represents the aggregation function in the $k$-th layer. HEC-GNN distinguishes itself from existing GNNs by mainly aggregating information from edge feature vectors. Moreover, HEC-GNN retains heterogeneity by modeling the interconnects between nodes with different relation types and separately gathering information of each relation type in the aggregation scheme.

Our HEC-GNN aggregation mechanism $\AGG^{(k)}$ is given as
\begin{equation}
	\label{eq:aggr}
	\footnotesize
	\AGG^{(k)}(\{ \boldsymbol{e}_{u,v,r}| r \in \mathcal{R}, u \in \mathcal{N}_v^r\}) = \smashoperator{\sum_{r\in \mathcal{R}}}\sum_{u\in \mathcal{N}_v^r}  {\boldsymbol{W}_r^{(k)} \boldsymbol{W}_\mathcal{E}^{(k)} \boldsymbol{e}_{u,v,r}},
\end{equation}
where $\boldsymbol{W}_\mathcal{E}^{(k)}$ and $\boldsymbol{W}_r^{(k)}$ are learnable weight matrices for all edges and the relation type $r\in\mathcal{R}$, respectively. Recall that in Eq.~\ref{eq:dyn}, the dynamic power can be viewed as weighted aggregation of interconnect activity $\alpha_i$ with weight being $C_iV^2f$. The heuristic behind Eq.~\ref{eq:aggr} is to simulate the formation of dynamic power consumption via the HEC-GNN aggregation and update process. Note that the activity $\alpha_i$ has already been encoded in the edge feature vector $\boldsymbol{e}_{u,v,r}$ with interconnect $i = (u,v,r)$, our goal is to adaptively fit the weight term $C_iV^2f$ via learnable weight matrices. To achieve this, we first use a global weight matrix $\boldsymbol{W}_\mathcal{E}^{(k)}$ to learn common knowledge from all types of edges, which corresponds to the equation term $V^2f$. Next, we simplify the intricate interconnects using multiple relation types. This enables us to approximate the interconnect capacitance $C_i$ using relation-specific interconnect capacitance $C_r$, which is produced by the weight matrix $\boldsymbol{W}_r^{(k)}$. Putting it all together, the edge feature vector $\boldsymbol{e}_{u,v,r}$ reflects the activity term $\alpha_i$, the global weight matrix $\boldsymbol{W}_\mathcal{E}^{(k)}$ accounts for $V^2f$, while the relation-specific weight matrices $\boldsymbol{W}_r^{(k)}, \forall r \in \mathcal{R}$, match with the interconnect capacitance $C_i$.
Hence, the HEC-GNN aggregation mechanism perceives dynamic power consumption by subtly fitting the dynamic power formula.

\setlength{\belowdisplayskip}{2pt}
\textbf{HEC-GNN overall architecture}. Fig.~\ref{fig:hecgnn} depicts the overall architecture of our HEC-GNN model for power learning. First, the graph data are fed into multiple HEC-GNN convolutional layers to learn and produce node embeddings. After that, the graph-level embedding $\boldsymbol{h}_G$ is obtained via sum pooling:
\begin{equation}
	\label{eq:pool}
	\small
	\boldsymbol{h}_G = \sum_{k\in \mathcal{K}}\sum_{v\in \mathcal{V}}{\boldsymbol{h}_v^{(k)}},
\end{equation}
where $\mathcal{K}$ is the set of indexes of graph convolutional layers. Instead of simply pooling the node embeddings from the last convolutional layer, we sum node embeddings obtained from different graph convolutional layers, which can be viewed as a form of skip connection to enhance generalization ability.

In addition to leveraging the graph embeddings that encapsulate localized information mainly contributing to dynamic power, we also exploit some global metedata from HLS reports that are complementary to the graph information and are indicative of static power. These metadata include the global resource utilization (LUT, DSP and BRAM), timing information (latency and achieved clock period) in HLS, and the scaling factors, i.e., the ratio of the above design metrics over those of the unoptimized baseline. As shown in Fig.~\ref{fig:hecgnn}, we use a multi-layer perceptron (MLP) with one fully connected layer followed by ReLU activation to embed the above global metedata. Thereafter, the two sets of embeddings, i.e., the graph embedding $\boldsymbol{h}_G$ and the global metadata embedding $\boldsymbol{h}_M$, are concatenated to form a holistic embedding. Finally, the holistic embedding is fed into another MLP with two fully connected layers and ReLU activation in between. The output of this MLP is the total or dynamic power estimation $P_{est.}$:
\begin{equation}
	\label{eq:mlp}
	\small
	P_{est.} = \MLP(\boldsymbol{h}_G\ ||\ \boldsymbol{h}_M).
\end{equation}

The above model components are cascaded to constitute our end-to-end supervised model, HEC-GNN, which is trained via regression to minimize the mean average percentage error loss. Furthermore, we adopt an ensemble learning strategy, in which we perform 10-fold cross-validation together with three different random seeds to generate different training and validation sets for model generation, and average all the output of trained models to get the final prediction results. 
Overall, HEC-GNN is heterogeneous, edge-centric and directed.

\begin{figure}[t]
	\begin{center}
		\includegraphics[width=\linewidth]{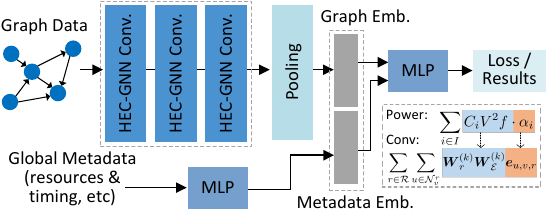}
		\vspace{-5mm}
		\caption{Overall architecture of HEC-GNN for power learning.}
		\label{fig:hecgnn}
	\end{center}
	\vspace{-8mm}
\end{figure}

\begin{table*}[t]
	\scriptsize
	\centering
	\vspace{-1mm}
	\caption{Dataset properties, results of total and dynamic power estimation, and runtime speedup over Vivado power estimator.}
	\vspace{-2mm}
	\label{table:acc}
	\begin{tabular}[width=\linewidth]{c|c|c||c|c|c||c|c|c|c|c|c||c}
		\toprule
		\multirow{2}{*}{Dataset} & \multicolumn{2}{c||}{Dataset Properties} & \multicolumn{3}{c||}{Error of Total Power (\%)} & \multicolumn{6}{c||}{Error of Dynamic Power (\%)} & \multirow{1}{*}{Runtime}\\
		& \multicolumn{1}{c|}{\#Samples} & \multicolumn{1}{c||}{Avg. \#Nodes} 
		& \multicolumn{1}{c|}{Vivado} & \multicolumn{1}{c|}{HL-Pow} & \multicolumn{1}{c||}{PowerGear} & 
		\multicolumn{1}{c|}{GCN} & \multicolumn{1}{c|}{GraphSage} & \multicolumn{1}{c|}{GraphConv} & \multicolumn{1}{c|}{GINE} & \multicolumn{1}{c|}{HL-Pow} & \multicolumn{1}{c||}{PowerGear} & \multicolumn{1}{c}{Speedup} \\
		\midrule
		\multicolumn{1}{c|}{Atax}    & 521 & 141 & 15.04 & \textbf{5.99} & 6.15 & 13.96 & 14.59 & 11.93 & 15.45 & 14.92 & \textbf{11.18} & 4.08$\times$  \\
		\multicolumn{1}{c|}{Bicg}    & 489 & 137 & 13.61 & \textbf{4.42} & 4.97 & 11.96 & 11.35 & 11.06 & 10.44 & 12.21 & \textbf{9.65}  & 3.96$\times$  \\
		\multicolumn{1}{c|}{Gemm}    & 531 & 341 & 25.59 & 3.22 & \textbf{2.75} & 11.02 & 9.66  & 9.53  & 8.92  & 10.35 & \textbf{8.32}  & 2.09$\times$  \\
		\multicolumn{1}{c|}{Gesummv} & 529 & 221 & 28.27 & 2.74 & \textbf{2.67} & 16.59 & 14.33 & 12.14 & 11.89 & 13.52 & \textbf{9.35}  & 10.81$\times$ \\
		\multicolumn{1}{c|}{2mm}     & 483 & 443 & 24.34 & \textbf{3.52} & 4.47 & 8.51  & 8.87  & 10.32 & 8.93  & 9.14  & \textbf{6.81}  & 1.99$\times$  \\
		\multicolumn{1}{c|}{3mm}     & 483 & 447 & 16.93 & \textbf{2.19} & 2.63 & 12.55 & 12.06 & 11.55 & 10.92 & 12.60 & \textbf{8.62}  & 2.31$\times$  \\
		\multicolumn{1}{c|}{Mvt}     & 531 & 165 & 19.03 & 2.98 & \textbf{2.77} & 13.94 & 12.09 & 9.94  & 10.04 & 12.38 & \textbf{8.77}  & 7.69$\times$  \\
		\multicolumn{1}{c|}{Syrk}    & 530 & 320 & 26.24 & 5.37 & \textbf{3.76} & 14.53 & 14.51 & 13.83 & 11.70 & 14.50 & \textbf{8.64}  & 1.47$\times$  \\
		\multicolumn{1}{c|}{Syr2k}   & 483 & 444 & 27.36 & 3.70 & \textbf{2.26} & 13.42 & 9.75  & 8.82  & 12.24 & 14.45 & \textbf{7.98}  & 2.13$\times$  \\
		\midrule
		\multicolumn{1}{c|}{Average} & 509 & 295 & 21.82 & 3.79 & \textbf{3.60} & 12.94 & 11.91 & 11.01 & 11.17 & 12.67 & \textbf{8.81}  & 4.06$\times$  \\
		\bottomrule
	\end{tabular}
	\vspace{-6mm}
\end{table*}

\section{Experimental Results}
The graph construction flow of PowerGear is implemented with Python for HLS data extraction and graph generation, and C++ for IR modification and detection probe implementation. The proposed HEC-GNN model and the baseline GNNs~\cite{kipf17,hamilton17,hu20,morris19} are developed with PyTorch Geometric~\cite{fey19} and Scikit-learn~\cite{Pedregosa11} toolkits. We evaluate our approach using Polybench~\cite{polybench} datasets with different dataset properties shown in Table~\ref{table:acc}. HLS design samples are generated by applying loop pipelining, loop unrolling and buffer partitioning. We also include some synthetic datasets to increase the diversity of loop patterns in training. The hardware designs are developed with Vivado and VivadoHLS 2018.2 and implemented on Xilinx Ultrascale+ ZCU102 FPGA board with the frequency of 100 MHz. Ground truth power values are obtained by power measurement with Power Advantage Tool~\cite{powtool}. Software design flows run on 80-core Intel Xeon CPU at 2.4 GHz with one Nvidia Tesla V100 GPU.	

Regarding the hyperparameters of GNNs, we employ three layers of graph convolution with a hidden dimension of 128, a batch size of 128, a dropout rate of 0.2 and a learning rate of 0.0005. We train the GNN models with 1200 and 2400 epochs for total and dynamic power estimation, respectively. 20\% of the training data are used as the validation set. 

\vspace{-1mm}
\subsection{Estimation Accuracy and Runtime Speedup}
In this experiment, we evaluate both total and dynamic power estimation accuracy and runtime speedup for HLS-based FPGA designs using PowerGear. We leave one target application out of the nine applications as the test dataset, and use all the others for training. With this leave-one-out training scheme, we can verify the transferability of the models on all nine datasets. We compare PowerGear with the Vivado power estimator~\cite{vpwr}, the state-of-the-art work HL-Pow~\cite{lin20}, and some mainstream GNN models~\cite{kipf17,hamilton17,hu20,morris19}. As for Vivado power estimation, we import the gate-level netlist after physical implementation and provide \textit{.saif} activity files via vector-based simulation, which ensures a high confidence level of estimation precision. Moreover, we observe through experiments that the Vivado power estimator neglects the impact of power gating~\cite{vpwrtech} on unused hard blocks, leading to a severe deviation from real power consumption. Hence, we further calibrate the results with a linear regression model. As for HL-Pow~\cite{lin20}, we follow its design flow and implement the gradient boosting decision tree (GBDT) models. Similar to GNNs, we use 20\% of training data for validation, based on which we tune the hyperparameters with tree size in [10, 500], tree depth in [5, 10], minimum samples per leaf in [2, 8], and learning rate in \{0.005, 0.01, 0.05\}.

As shown in Table~\ref{table:acc}, regarding total power estimation, Vivado power estimation diverges considerably from the real measurement, leading to an average error of 21.82\%, whereas the HL-Pow and our proposed PowerGear achieve good prediction accuracy with average errors of 3.79\% and 3.60\%, respectively. In general, PowerGear achieves the best average accuracy for total power estimation. Regarding dynamic power estimation, we compare PowerGear with HL-Pow and commonly used GNN models~\cite{kipf17,hamilton17,hu20,morris19}. Overall, PowerGear outperforms all baseline methods on all the evaluated datasets, significantly reducing the estimation error of dynamic power from 12.67\% down to 8.81\% compared with HL-Pow. Results verify that PowerGear's graph construction flow successfully retains interconnect and activity features that have important implication for power consumption, and moreover, the HEC-GNN model proposed in PowerGear effectively mines rich heterogeneous edge semantics and structural properties via the edge-centric neighborhood aggregation mechanism. In comparison to the GNN baselines, including GCN~\cite{kipf17} and GraphSage~\cite{hamilton17} that simply focus on node features, and GraphConv~\cite{morris19} and GINE~\cite{hu20} which use both node and edge features, PowerGear still reaps notable accuracy improvement. Our insight is that PowerGear intelligently fits the dynamic power formulation in HEC-GNN aggregation and additionally accounts for overall power characteristics using metadata embeddings, leading to high efficacy and generalization ability for power modeling.

With PowerGear, we can realize efficient FPGA power estimation by simply executing the HLS flow and skipping the tedious cycle-accurate simulation, low-level synthesis and physical implementation steps. This gives rise to the runtime speedup of 1.47--10.81$\times$ (4.06$\times$ on average) over the Vivado power estimation process.

\vspace{-1.5mm}
\subsection{Ablation Study}
\begin{table}[t]
	\centering
	\vspace{1mm}
	\caption{Error (\%) of dynamic power estimation using different HEC-GNN variants.}
	\vspace{-2mm}
	\label{table:abl}
	\resizebox{\columnwidth}{!}{
		\begin{tabular}[width=\linewidth]{c||c|c|c|c|c|c||c}
			\toprule
			\multicolumn{1}{c||}{Dataset} & \multicolumn{1}{c|}{W/o opt.} & \multicolumn{1}{c|}{W/o e.f.} & \multicolumn{1}{c|}{W/o dir.} & \multicolumn{1}{c|}{W/o hetr.} & \multicolumn{1}{c|}{W/o md.} & \multicolumn{1}{c||}{Sgl.} & \multicolumn{1}{c}{Prop.}\\
			\midrule
			\multicolumn{1}{c||}{Atax}    & 13.32 & 11.40 & 11.84 & 11.39 & 12.37 & 11.68 & \textbf{11.18} \\
			\multicolumn{1}{c||}{Bicg}    & 10.63 & 10.56 & 10.19 & 9.98  & 11.08 & 10.06 & \textbf{9.65} \\
			\multicolumn{1}{c||}{Gemm}    & 10.04 & 10.39 & 8.53  & 9.48  & 8.66  & 8.79  & \textbf{8.32} \\
			\multicolumn{1}{c||}{Gesummv} & 13.83 & 11.14 & 9.99  & 10.31 & 9.74  & 9.68  & \textbf{9.35} \\
			\multicolumn{1}{c||}{2mm}     & 10.25 & 8.67 & 6.97  & 7.64  & 7.47  & 6.93  & \textbf{6.81} \\
			\multicolumn{1}{c||}{3mm}     & 11.77 & 9.62 & 9.29  & 9.15  & 11.93 & 8.96  & \textbf{8.62} \\
			\multicolumn{1}{c||}{Mvt}     & 12.54 & 9.40 & 8.70  & 9.71  & 8.81  & \textbf{8.47} & 8.77 \\
			\multicolumn{1}{c||}{Syrk}    & 12.64 & 10.92 & 8.95  & 9.51  & 10.75 & 9.04  & \textbf{8.64} \\
			\multicolumn{1}{c||}{Syr2k}   & 9.62 & 9.66 & 8.51  & 8.97  & \textbf{7.16}  & 8.10  & 7.98 \\
			\midrule
			\multicolumn{1}{c||}{Average} & 11.74 & 10.20 & 9.22  & 9.57  & 9.77  & 9.08  & \textbf{8.81} \\
			\bottomrule
		\end{tabular}
	}
	\vspace{-6mm}
\end{table}
To gain better understanding of various optimization strategies adopted by HEC-GNN in PowerGear, we conduct an ablation study which evaluates the accuracy of dynamic power estimation using six variants of HEC-GNN in addition to the proposed HEC-GNN denoted as \textit{prop.} for short. These models include: 1) \textit{w/o opt.}: single unoptimized HEC-GNN without considering edge features, directionality, heterogeneity, and global metadata embeddings; 2) \textit{w/o e.f.}: single HEC-GNN without using edge features; 3) \textit{w/o dir.}: single HEC-GNN without directionality; 4) \textit{w/o hetr.}: single HEC-GNN without using multiple relation types; 5) \textit{w/o md.}: single HEC-GNN without using the global metadata embeddings; and 6) \textit{sgl.}: single optimized HEC-GNN without considering ensemble. 

As shown in Table~\ref{table:abl}, our proposed HEC-GNN model is consistently superior to its six variants in seven out of nine datasets for dynamic power estimation with FPGA HLS, and finally yields the best accuracy on average. It can be inferred from the results that the adoption of edge features, directionality, heterogeneity, global metadata embeddings and ensemble all contribute to the enhancement of model accuracy, which explains the effectiveness of HEC-GNN that jointly makes use of all these optimization strategies.

\vspace{-1mm}
\subsection{Case Study}
\begin{figure}[t]
	\begin{center}
		\begin{subfigure}[b]{0.47\linewidth}
			\includegraphics[width=\linewidth]{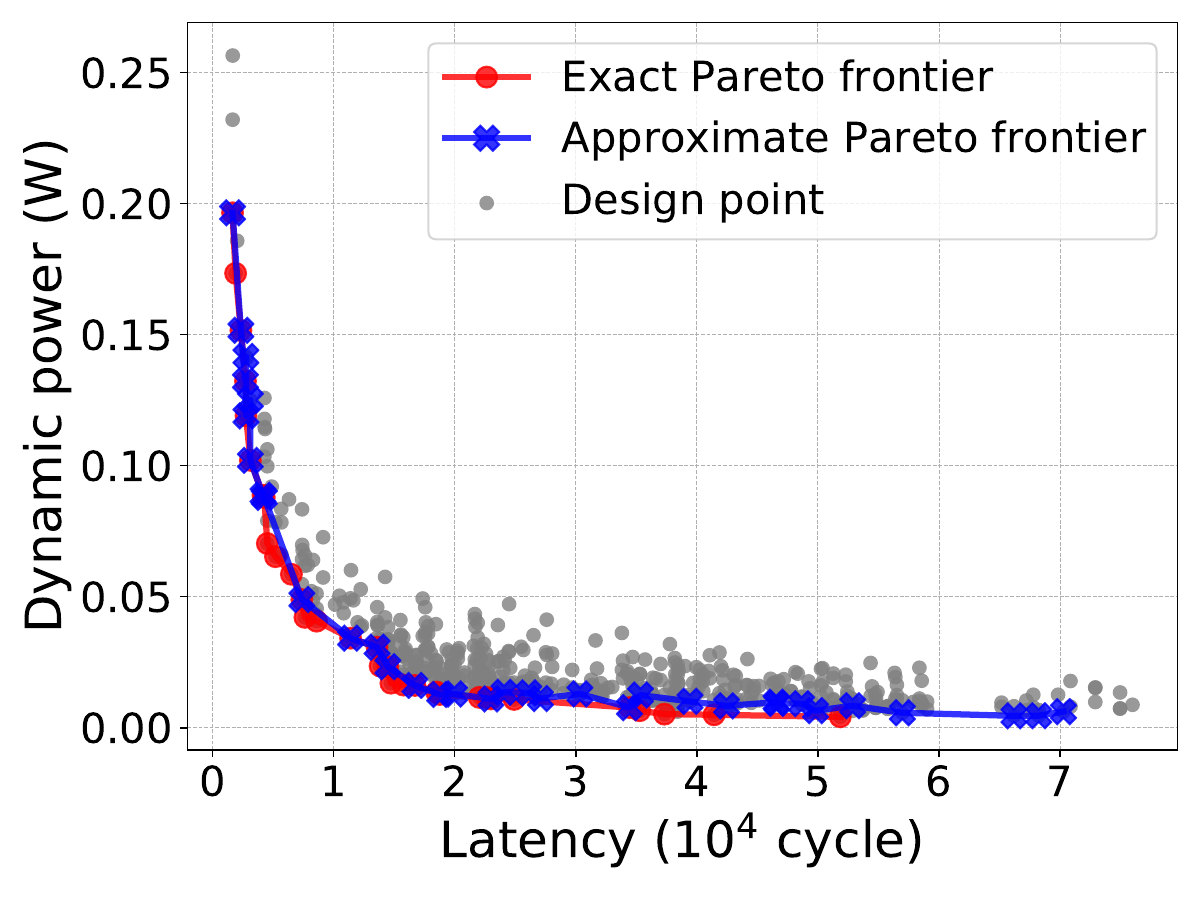}
			\vspace{-6mm}
			\caption{Atax}
		\end{subfigure}
		\begin{subfigure}[b]{0.47\linewidth}
			\includegraphics[width=\linewidth]{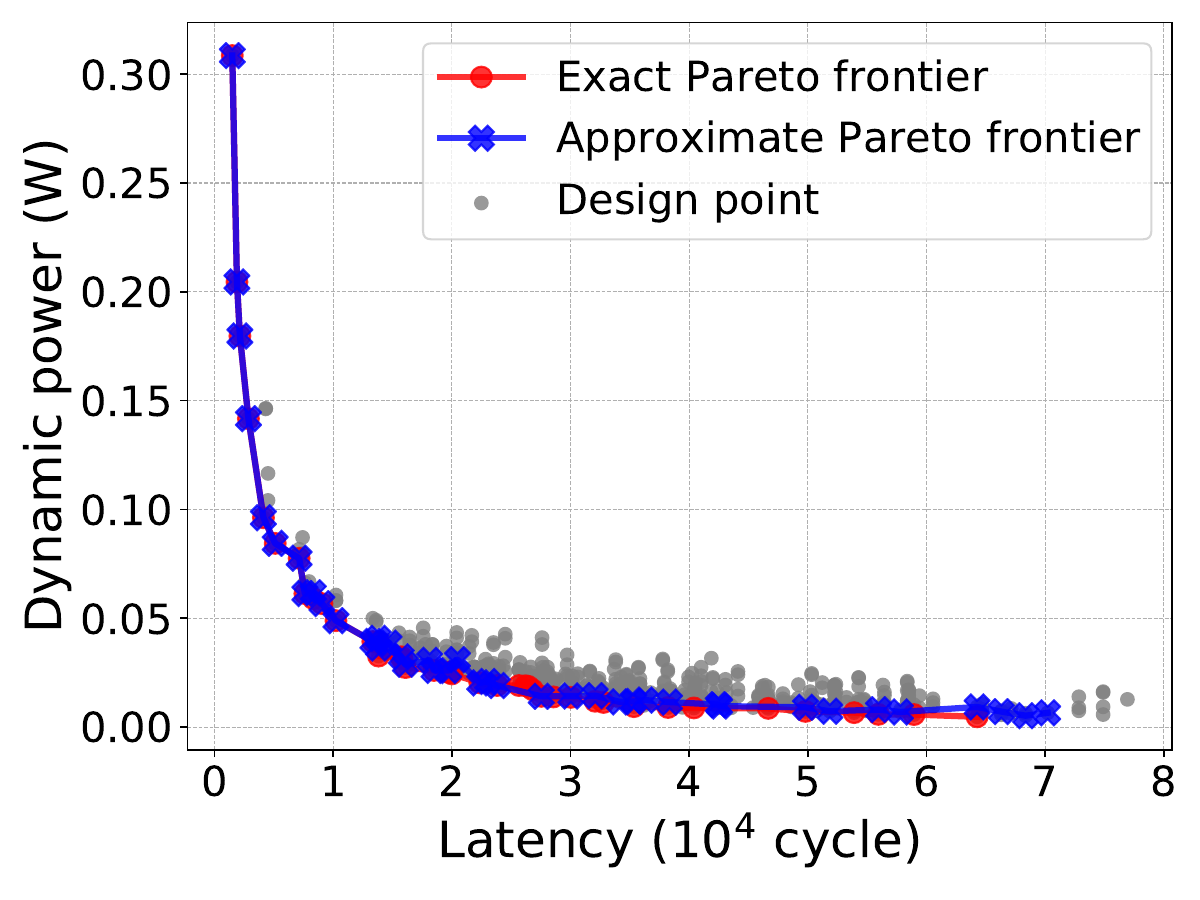}
			\vspace{-6mm}
			\caption{Mvt}
		\end{subfigure}
		\vspace{-2mm}
		\caption{Pareto frontiers of Atax and Mvt with PowerGear.}
		\label{fig:pareto}
	\end{center}
	\vspace{-4mm}
\end{figure}

\begin{table}[t]
	\scriptsize
	\centering
	\vspace{-1mm}
	\caption{ADRS of HLS-based design space exploration.}
	\vspace{-2mm}
	\label{table:adrs}
	\begin{tabular}[width=\linewidth]{c||c|c|c||c|c}
		\toprule
		\multicolumn{1}{c||}{Sampling} & \multicolumn{3}{c||}{Prediction Model} & \multicolumn{2}{c}{PowerGear Gains}\\
		\multicolumn{1}{c||}{Budget} & \multicolumn{1}{c|}{Vivado} & \multicolumn{1}{c|}{HL-Pow} & \multicolumn{1}{c||}{PowerGear} & \multicolumn{1}{c|}{v.s. Vivado} & \multicolumn{1}{c}{v.s. HL-Pow}\\
		\midrule
		\multicolumn{1}{c||}{20\%}  & 0.1657 & 0.1050 & 0.0981 & 39.2\% & 6.9\%  \\
		\multicolumn{1}{c||}{30\%}  & 0.1520 & 0.0841 & 0.0774 & 45.8\% & 10.4\% \\
		\multicolumn{1}{c||}{40\%}  & 0.1423 & 0.0691 & 0.0626 & 52.0\% & 11.2\% \\
		\bottomrule
	\end{tabular}
	\vspace{-6mm}
\end{table}

We demonstrate a case study in which PowerGear is exploited as the prediction model to guide fast HLS-based hardware design space exploration (DSE) to trade off between latency and dynamic power. Given a dataset for DSE, we first sample a small subset of design points for HLS and then utilize PowerGear to estimate dynamic power. Together with the set of latency derived from HLS, we compute the dynamic power-latency Pareto frontier using existing sampling points, based on which a sampling algorithm~\cite{lin20} is applied to select promising design points that are most likely to be Pareto-optimal for further evaluation. The above steps are conducted iteratively to calibrate the approximate Pareto frontier until the total sampling budget is met. Experiments are performed with an initial sampling budget of 2\% and total sampling budgets of 20\%, 30\% and 40\%, respectively. Vivado power estimator and HL-Pow are used as alternative power prediction models for comparison. A commonly used evaluation metric \textit{average distance from reference set (ADRS)} is employed to quantify the difference between the approximate and exact Pareto frontiers:
\begin{equation}
	\label{eq:adrs}
	\small
	\ADRS(\Gamma,\Omega) = \frac{1}{|\Gamma|}\sum_{\gamma\in\Gamma}\min_{\omega\in\Omega}f(\gamma,\omega),
\end{equation}
where $\Omega$ is the approximate Pareto-optimal set, $\Gamma$ is the exact Pareto-optimal set, and $f(\cdot)$ computes the distance between two design points $\omega\in\Omega$ and $\gamma\in\Gamma$. A lower ADRS means that the approximate Pareto set is closer to the exact one. 

Fig.~\ref{fig:pareto} shows the DSE results of two datasets under a total sampling budget of 40\%, which indicate that PowerGear can effectively aid in DSE algorithms to realize close approximation of Pareto-optimal sets with moderate runtime overhead. Moreover, as shown in Table~\ref{table:adrs}, the ADRS of Pareto search using PowerGear as the prediction model is the best for all three sampling budgets. Compared with the methods using Vivado and HL-Pow, PowerGear-assisted DSE yields relative performance gains of 39.2--52\% and 6.9--11.2\%, respectively. Through this case study, we show that PowerGear's capability to provide accurate dynamic power estimation in FPGA HLS opens up more opportunities for optimizing dynamic power.
	
	\section{Conclusion}
	We propose PowerGear, an early-stage power estimation approach for FPGA HLS. PowerGear constructs graph data from HLS in a way that preserves operations, interconnects and switching activities, and moreover, PowerGear subtly exploits edge features for power learning via HEC-GNN. Experiments prove that PowerGear is accurate, efficient and transferable, and can notably boost the quality of DSE in FPGA HLS.
	
	\section*{Acknowledgment}
	This research was supported by Key-Area Research and Development Program of Guangdong Province under grant No. 2019B010155002.
	
	\bibliographystyle{IEEEtran}
	\footnotesize{
		\bibliography{ref}
	}
	
\end{document}